%% file: main.tex
\documentclass[10pt,twocolumn,letterpaper]{article}

\usepackage[pagenumbers]{iccv} 

\input{preamble}

\usepackage[pagebackref,breaklinks,colorlinks,allcolors=iccvblue]{hyperref}
\usepackage[table,dvipsnames]{xcolor}
\usepackage{colortbl}
\definecolor{iccvblue}{rgb}{0.21,0.49,0.74}
\usepackage{siunitx}
\def\paperID{5879} 
\def\confName{ICCV}
\def\confYear{2025}

\title{Training-Free Text-Guided Image Editing with Visual Autoregressive Model}

\author{
{Yufei Wang$^{1,2}$,
Lanqing Guo$^3$, 
Zhihao Li$^{2   }$, 
Jiaxing Huang$^2$, 
Pichao Wang,
Bihan Wen$^2$, 
Jian Wang$^1$} \vspace{2mm}
\\
{$^1$Snap Research \quad $^2$Nanyang Technological University \quad $^3$ UT Austin\vspace{3mm}} 
\\
\textit{Code will be at \url{https://github.com/wyf0912/AREdit}} 
}
\begin{document}

\twocolumn[{%
\renewcommand\twocolumn[1][]{#1}%
\maketitle
\includegraphics[width=\linewidth]{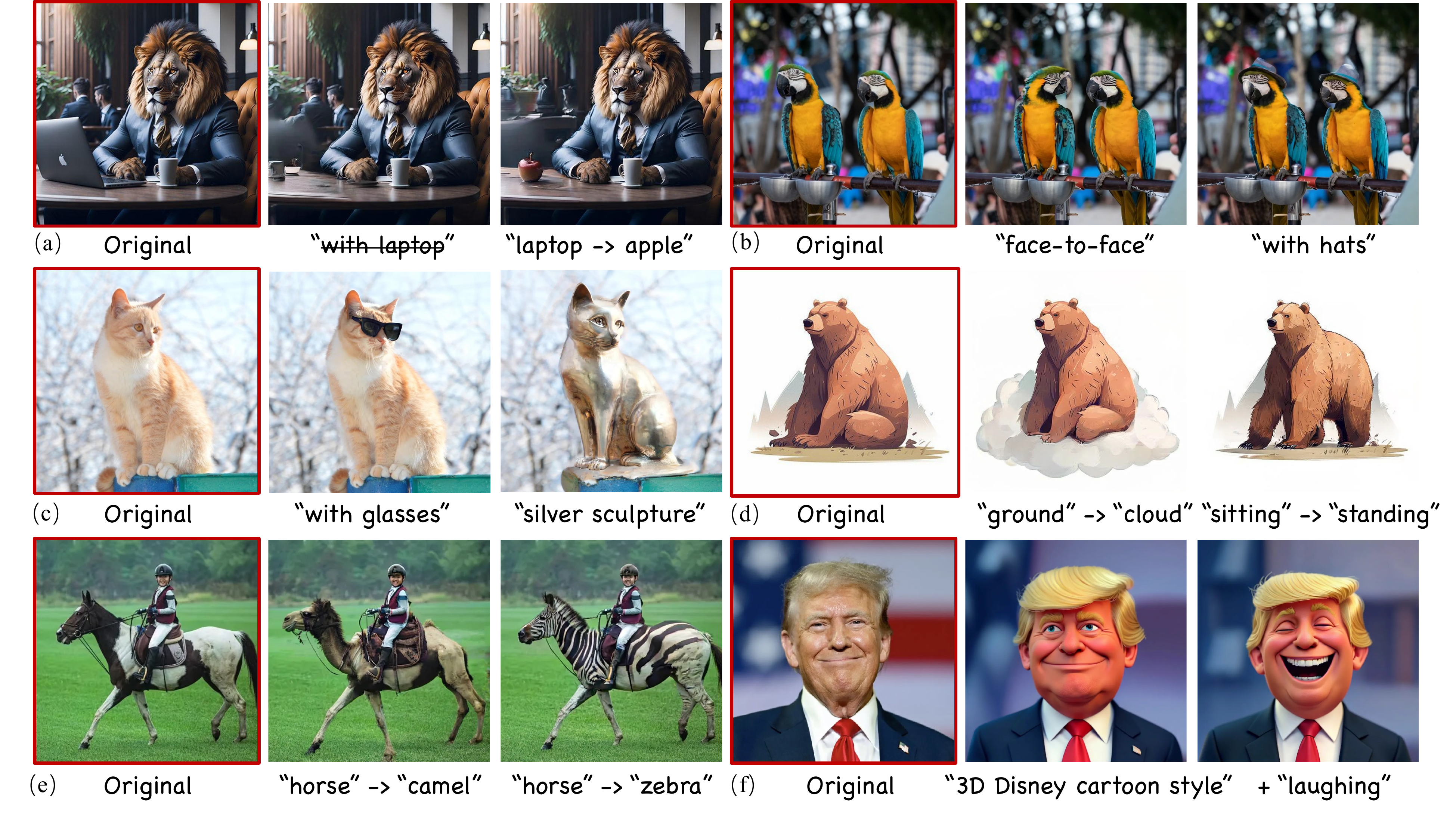}
\captionof{figure}{\textbf{AREdit for Text-Guided Image Editing.} It can effectively handle a variety of editing tasks for both artificial and natural images, \eg, object removal (as in examples a), object addition (b, c), attribute modification (b, d), and style alteration (c, f). Our method excels at preserving unrelated areas of the image while offering a flexible trade-off with generative capacity. Remarkably fast, our approach processes a \textbf{1080p input image} in 2.5 seconds for the first run and \textbf{$\sim$1.2 second }for subsequent runs on an A100 GPU.
\vspace{1em}}
\label{fig:teaser}
}]

\input{sec/0_abstract}    
\input{sec/1_intro}

\input{sec/2_related_work}

\input{sec/3_method}
\input{sec/4_exp}
\input{sec/5_conclusion}
{
    \small
    \bibliographystyle{ieeenat_fullname}
    \bibliography{main}
}

\end{document}

%% file: preamble.tex
\usepackage{algorithm}
\usepackage{algpseudocode}
\usepackage{bbm}
\newcommand{\myComment}[1]{\hfill\small\textcolor{gray}{$\triangleright$ \textit{#1}}}
\newcommand{\R}[0]{\mathbf{R}}
\newcommand{\F}[0]{\mathbf{F}}
\newcommand{\I}[0]{\mathbf{I}}
\newcommand{\C}[0]{\mathbf{C}}
\renewcommand{\P}[0]{\mathbf{P}}
\newcommand{\M}[0]{\mathbf{M}}
\newcommand{\up}[0]{\text{up}}
\newcommand{\W}[0]{\mathbf{W}}

%% file: sec/0_abstract.tex
\begin{abstract}
Text-guided image editing is an essential task that enables users to modify images through natural language descriptions. 
Recent advances in diffusion models and rectified flows have significantly improved editing quality, primarily relying on inversion techniques to extract structured noise from input images. However, inaccuracies in inversion can propagate errors, leading to unintended modifications and compromising fidelity. Moreover, even with perfect inversion, the entanglement between textual prompts and image features often results in global changes when only local edits are intended.
To address these challenges, we propose a noel text-guided image editing framework based on VAR (Visual AutoRegressive modeling), which eliminates the need for explicit inversion while ensuring precise and controlled modifications. 
Our method introduces a caching mechanism that stores token indices and probability distributions from the original image, capturing the relationship between the source prompt and the image. 
Using this cache, we design an adaptive fine-grained masking strategy that dynamically identifies and constrains modifications to relevant regions, preventing unintended changes.
A token re-assembling approach further refines the editing process, enhancing diversity, fidelity and control. 
Our framework operates in a training-free manner and achieves high-fidelity editing with faster inference speeds, processing a 1K resolution image in as fast as 1.2 seconds. 
Extensive experiments demonstrate that our method achieves performance comparable to, or even surpassing, existing diffusion- and rectified flow-based approaches in both quantitative metrics and visual quality. The code will be released.
\end{abstract}

%% file: sec/1_intro.tex
\section{Introduction}
\label{sec:intro}
With the rapid advancement of generative AI, text-guided image editing~\cite{meng2021sdedit,brooks2023instructpix2pix,kawar2023imagic,hertz2022prompt,balaji2022ediff,xu2023inversion,feng2024dit4edit} has emerged as a crucial technique in computer vision. By enabling users to modify images through natural language descriptions, it makes image manipulation more intuitive and accessible. This technology has a wide range of applications, including artistic content creation, automated image enhancement, and interactive design.

Recent progress in generative models, particularly diffusion (DF) models~\cite{ho2020denoising,ldm,vqdm,sdxl} and rectified flows (RFs)~\cite{liu2022flow,flux2024}, has led to significant improvements in automated image editing. A common approach involves using inversion techniques~\cite{song2020denoising,gal2022image,xu2023inversion,ju2023direct} to extract structured noise from an input image, which serves as an initialization for modifications based on a given prompt. This method enables edited results that maintain the original structure while incorporating the specified changes. However, despite their effectiveness, existing techniques still struggle with fidelity preservation, where unintended modifications often appear in regions that should remain unchanged.

The unintended changes from inversion-based methods primarily arise from two key issues. The first challenge is achieving accurate inversion. Current methods~\cite{meng2021sdedit,balaji2022ediff} rely on predicting structured noise from an input image and its source prompt, but errors in this process can propagate through the editing pipeline, leading to visual artifacts or unintended semantic changes. The second challenge is the entanglement between image features and textual prompts. Even with perfect inversion, the interplay between text and image features, combined with accumulated errors in generative models, often results in local edits affecting the entire image in undesirable ways. This limits the fine-grained controllability required for precise editing.

Several approaches have been proposed to address these issues~\cite{ju2023direct, mokady2023null, wang2024taming}. For example, RF-Solver~\cite{wang2024taming} applies high-order Taylor expansion to approximate nonlinear components in the inversion process, improving noise reconstruction accuracy. However, residual errors remain, and the additional computational overhead makes the approach inefficient. Despite these efforts, there is still a need for an efficient and robust framework that ensures precise modifications while maintaining fidelity in unedited regions.

To address these limitations, we propose \textit{AREdit}, a novel text-guided image editing framework based on VAR that enables high-fidelity modifications without additional training. Unlike existing approaches that rely on iterative noise refinement, this framework adopts a next-scale prediction strategy, progressively generating higher-resolution features from a coarse-to-fine manner. This approach allows for efficient and structured modifications without the need for diffusion-based sampling. Our method introduces three key components. 
First, we develop a cache mechanism that stores token indices and probability distributions from the original image, providing important clues about the information gap between the given text prompt and the input image. This cached information serves a role similar to structured noise in diffusion models and rectified flows but offers greater flexibility. 
Second, we introduce an adaptive fine-grained masking strategy that dynamically identifies the regions requiring modification by computing the distributional differences between the original and edited prompts. This ensures that only relevant regions are modified while preserving unedited content. 
Third, we employ a low-frequency-preserving token re-assembling approach that reuses low-frequency features while refining the subsequent sampling process to generate tokens aligned with the desired edit, enhancing the precision, diversity and control of the modifications.

Our contributions can be summarized as follows.
\begin{itemize}
    \item We present the first VAR-based text-guided image editing framework that enables high-quality, training-free modifications by exploiting token distribution differences between the randomness caching and editing process.
    \item We propose an adaptive fine-grained masking and token re-assembling strategy, compatible with existing attention control methods, which ensures precise local editing, promotes diversity across multiple samplings, and maintains global consistency.
    \item Our method achieves performance comparable to, or even surpassing, existing diffusion and rectified flow-based approaches across multiple metrics, delivering visually superior editing results, and is $\sim$9$\times$ faster than existing SOTA methods based on diffusion/rectified flows.
\end{itemize}

%% file: sec/2_related_work.tex
\begin{figure*}[t]
    \centering
    \includegraphics[width=1\linewidth]{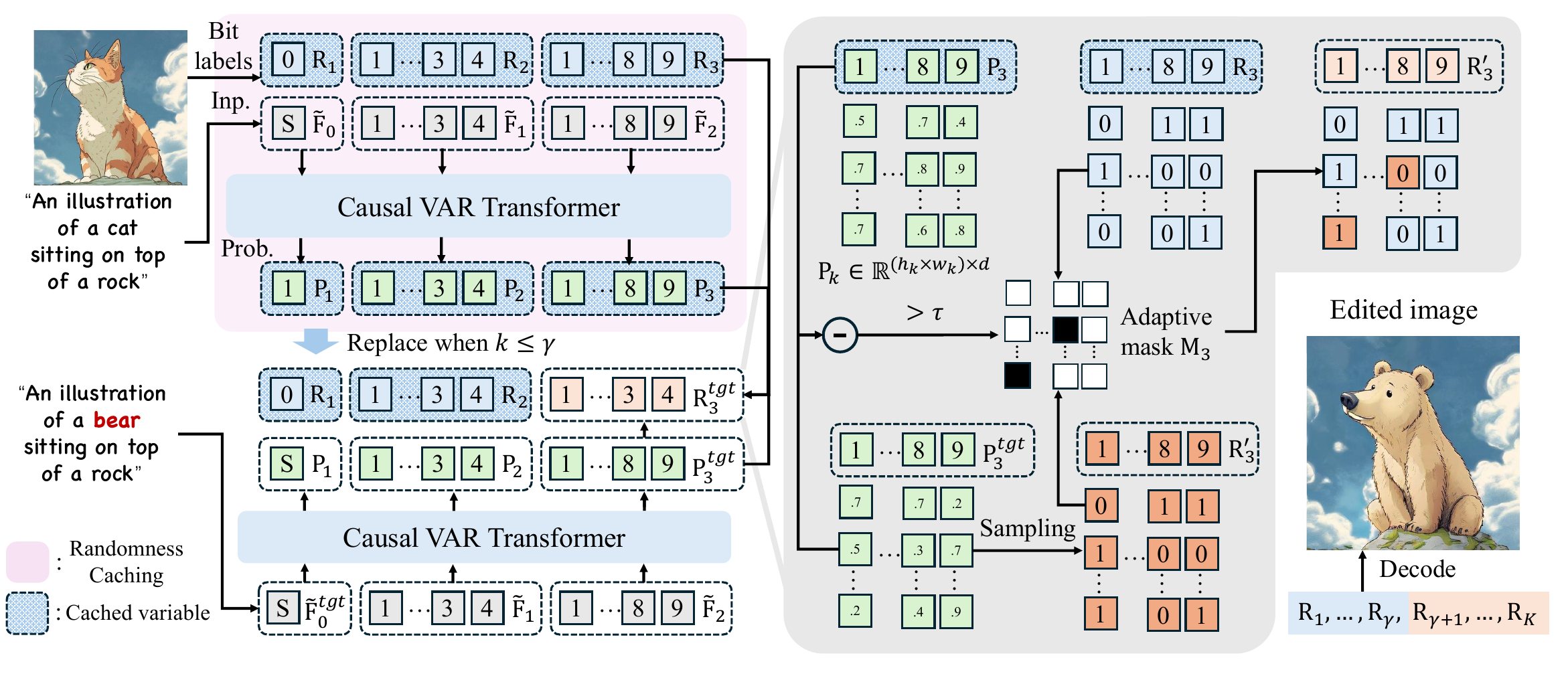}
\vspace{-0.5cm}
\caption{The overall framework of the proposed method built on the pretrained Infinity~\cite{han2024infinity}. Given an input image and its text prompt, we first cache the bit labels $\R_{queue}$ and probability distributions $\P_{queue}$ for editing using a feedforward evaluation as in Algorithm~\ref{alg:caching}. During editing, for steps where $k \leq \gamma$, cached bit labels are reused to preserve the overall appearance and structure of the image. For steps where $k > \gamma$, we compute a fine-grained adaptive mask, $\M_k$, based on the probability distributions $\P_k$ and $\P_k^{tgt}$. Here, $\P_k$ is from the cached distributions $\P_{cache}$, and $\P_k^{tgt}$ is predicted by conditioning on the target prompt, $t_T$. The fine-grained mask $\M_k \in \mathbb{R}^{(h_k \times w_k) \times d}$ is then employed to blend the cached bit labels $\R_k$ with the randomly sampled ones $\R_k'$ from the distribution $\P_k^{tgt}$. Finally, the decoder generates the edited image using the content $(\R_1, ..., \R_\gamma, \R^{tgt}_{k+1}, ..., \R^{tgt}_{K})$. For illustration purposes, we use $K=3$ and $\gamma=2$ as an example.}
\label{fig:framework}
\vspace{-0.2cm}
\end{figure*}

\section{Related works}
\label{sec:related_works}
\subsection{Text-guided image editing}
With the rise of text-guided diffusion models~\cite{rombach2022high,ramesh2022hierarchical}, which naturally align the image latents and text embeddings, a series of training-free editing methods~\cite{meng2021sdedit,kawar2023imagic,hertz2022prompt,balaji2022ediff, xu2023inversion} have emerged. These methods eliminate the need for training data while preserving the naturalness and high quality of the edited images to the greatest extent.
Most of these methods use inversion techniques~\cite{song2020denoising,dhariwal2021diffusion}, where a structured noise is obtained by inverting the image, and a new prompt is then fed into the T2I generative model.
For example, some methods utilize latent blending techniques, such as SDEdit~\cite{meng2021sdedit} and \cite{avrahami2022blended}, which inject editing instructions through noisy latent features. Additionally, attention-sharing mechanisms, like Prompt-to-Prompt~\cite{hertz2022prompt} and plug-and-play (PnP)~\cite{tumanyan2023plug}, are commonly employed to fuse the attention maps connecting text and image. Furthermore, some approaches focus on improved inversion techniques~\cite{ju2023direct, lin2025schedule} to improve reconstruction quality and editing perception.  For instance, Ju~\etal~\cite{ju2023direct} propose a direct inversion approach that disentangles responsibilities, ensuring both essential content preservation and edit fidelity.
More recently, Rectified Flow (RF) models, such as Flux~\cite{flux2024}, have shown promising results. Building on this, subsequent works on RF inversion and editing~\cite{rout2025semantic, wang2024taming, zheng2024oscillation} have been proposed.
However, diffusion-based methods, including rectified flows, still require long processing times and struggle to preserve original structures due to the necessary inversion, which significantly limit their practical applicability.
In this paper, we propose the first VAR based text-guided image editing algorithm, enabling high-fidelity editing with signficantly faster speed.

\subsection{Visual autoregressive modeling}
Inspired by the strong scaling capabilities of large language models (LLMs), autoregressive (AR) models in vision represent images as sequences of discrete tokens or pixels and model their dependencies using architectures such as transformers.
Earlier works like VQVAE~\cite{van2017neural} and VQGAN~\cite{esser2021taming} applied vector quantization to encode image patches as indexed tokens, using a decoder-only transformer for next-token prediction. However, the lack of large-scale transformers and inherent quantization errors have limited their performance, preventing them from matching diffusion models.
Some recent efforts~\cite{wang2024emu3,kondratyuk2023videopoet,yu2022scaling} are demonstrated that scaling up AR models improved the image and video generation performances.
Building on the global structure of visual information, VAR~\cite{VAR} introduces a novel visual sequence based on next-scale prediction, significantly enhancing generation quality and sampling efficiency.
To advance AR models for text-guided image generation, HART~\cite{tang2024hart} introduces a hybrid tokenizer that integrates both continuous and discrete tokens, complemented by a lightweight diffusion model for residual learning.
Infinity~\cite{Infinity} redefines the framework as bitwise token prediction, incorporating an infinite-vocabulary tokenizer, a classifier, and a bitwise self-correction mechanism.
With the emergence of powerful foundation models, training-free AR-based text-guided editing has become possible.

%% file: sec/3_method.tex
\section{Methodology}
\subsection{Preliminary}
Visual AutoRegressive Modeling (VAR)~\cite{tian2025visual} includes two main parts: a visual tokenizer (consisting of an encoder $\mathcal{E}$ and a decoder $\mathcal{D}$) for encoding and decoding image and a transformer~$\mathcal{T}$ for image synthesis. Specifically, the visual tokenizer first encodes the image $\I$ into a continuous feature map $\F \in \mathbb{R}^{h \times w \times d}$ and then quantizes it into $K$ multi-scale discrete residual maps $(\R_1, \R_2, ..., \R_K)$. Using this sequence of residuals, we can progressively approximate the continuous feature $\F$ as:
\begin{equation}
    \F_k = \sum_{i=1}^k \up(\R_i, (h, w)),
\label{eq:var_1}
\end{equation}
where $\up(\R_i, (h, w))$ represents the operation that upsample the feature $\R_i$ to the resolution of $(h, w)$. The transformers predict the discrete residual of current level $\R_k$ based on the context of all previous levels, \ie, $(\R_1, ..., \R_{k-1})$, and the text embeddings $\Psi(t)$ from Flan-T5~\cite{chung2024scaling} in an autoregressive manner as follows
\begin{equation}
    p(\R_1, ..., \R_K) = \prod_{k=1}^{K} p(\R_k | \R_1, ..., \R_{k-1}, \Psi(t)).
\end{equation}
Specifically, for each step that predicts $\R_k$ based on its context, a bilinear downsampled feature $\tilde{\F}_{k-1}$ is used as the input. This feature is obtained by downsampling $\F_{k-1}$ to the dimensions $(h_k, w_k)$ using bilinear interpolation, where the input of the first scale $\tilde{\F}_0$ is set to $\text{⟨SOS⟩}$ which is projected from the text embedding~\cite{han2024infinity}.

We adopt Infinity-2B~\cite{han2024infinity} as the backbone of our work, which is a recent advent based on VAR~\cite{tian2025visual}. Different from the original vector quantizer of VAR which leads to unaffordable space and time complexity of $O(2^d)$ when using a large vocabulary dimension $d$, Infinity adapts BSQ quantizer~\cite{zhao2024image} that maps the continuous residual vector $z_k\in \mathbb{R}^d$ into discrete format $q_k$ as follows
\begin{equation}
    q_k = \frac{1}{\sqrt{d}} \operatorname{sign}\left(\frac{z_{k}}{|z_{k}|}\right).
\end{equation}
Correspondingly, $d$ binary classifiers are used in parallel to predict the next-scale residual $\R_k$ is positive or negative, which is more efficient in memory and parameters compared with the vanilla one.

\begin{algorithm}[t]
\caption{Randomness caching}
\begin{algorithmic}[1]
\Require Image $\I$ to edit, encoder $\mathcal{E}$, transformer $\mathcal{T}$, and text embeddings $\Phi(t_S)$ from the source prompt $t_S$
\State $\tilde{\F}_{queue}, \R_{queue} = \mathcal{E}(\I)$
\myComment {$R_{queue}$: bit labels} \\
\myComment {$\tilde{\F}_{queue}$: inputs for transformer}
\State $\P_{queue} = \W_{queue} = <>$ \myComment{empty queue for caching}
\State $\tilde{\F}_0 = \text{⟨SOS⟩} \in \mathbb{R}^
{1\times1\times h}$ \myComment{{\normalfont⟨SOS⟩} is the start token~\cite{han2024infinity}}
\For{$k = 1, 2, \dots, K$}
\State $\tilde{\F}_k$ = \Call{Queue\_Pop}{$\tilde{\F}_{queue}$}
\State $\P_{k}, \W_k = \mathcal{T}(\tilde{\F}_{k-1}, \Phi(t_S))$ \myComment {probability distributions}
\State \Call{Queue\_Push}{$\P_{queue}, \P_{k}$}
\State \Call{Queue\_Push}{$\W_{queue}, \W_{k}$} \myComment{for attention control}
\EndFor
\State \textbf{Output:} $\R_{queue}, \P_{queue}, \W_{queue}$
\end{algorithmic}
\label{alg:caching}
\end{algorithm}

\begin{table*}[t]
    \centering
\begin{tabular}{l|c|c|ccc|cc}
\hline
\textbf{Method} & \textbf{Base} & \textbf{Structure} & \multicolumn{3}{c|}{\textbf{Background Preservation}} & \multicolumn{2}{c}{\textbf{CLIP Similarity}↑} \\
 &\textbf{Model} & Distance↓ & PSNR↑ & SSIM↑ & LPIPS↓ & Whole & Edited \\
\hline
Prompt2Prompt~\cite{hertz2022prompt} & diffusion & \cellcolor[rgb]{1,1.000,0.8} 0.0694 & \cellcolor[rgb]{1,1.000,0.8} 17.87 & \cellcolor[rgb]{1,1.000,0.8} 0.7114 & \cellcolor[rgb]{1,1.000,0.8} 0.2088 & \cellcolor[rgb]{1,0.878,0.8} 25.01 & \cellcolor[rgb]{1,0.866,0.8} 22.44 \\
Pix2Pix-Zero~\cite{parmar2023zero}& diffusion & \cellcolor[rgb]{1,0.966,0.8} 0.0617 & \cellcolor[rgb]{1,0.919,0.8} 20.44 & \cellcolor[rgb]{1,0.944,0.8} 0.7467 & \cellcolor[rgb]{1,0.940,0.8} 0.1722 & \cellcolor[rgb]{1,1.000,0.8} 22.80 & \cellcolor[rgb]{1,1.000,0.8} 20.54 \\
MasaCtrl~\cite{cao2023masactrl}& diffusion & \cellcolor[rgb]{1,0.818,0.8} 0.0284 & \cellcolor[rgb]{1,0.864,0.8} 22.17 & \cellcolor[rgb]{1,0.864,0.8} 0.7967 & \cellcolor[rgb]{1,0.832,0.8} 0.1066 & \cellcolor[rgb]{1,0.936,0.8} 23.96 & \cellcolor[rgb]{1,0.956,0.8} 21.16 \\
PnP~\cite{tumanyan2023plug}& diffusion & \cellcolor[rgb]{1,0.817,0.8} 0.0282 & \cellcolor[rgb]{1,0.860,0.8} 22.28 & \cellcolor[rgb]{1,0.874,0.8} 0.7905 & \cellcolor[rgb]{1,0.843,0.8} 0.1134 & \cellcolor[rgb]{1,0.856,0.8} 25.41 & \cellcolor[rgb]{1,0.858,0.8} 22.55 \\
PnP-DirInv.~\cite{ju2023direct}& diffusion & \cellcolor[rgb]{1,0.800,0.8} 0.0243 & \cellcolor[rgb]{1,0.855,0.8} 22.46 & \cellcolor[rgb]{1,0.864,0.8} 0.7968 & \cellcolor[rgb]{1,0.831,0.8} 0.1061 & \cellcolor[rgb]{1,0.856,0.8} 25.41 & \cellcolor[rgb]{1,0.853,0.8} 22.62 \\
LEDits++~\cite{brack2024ledits}& diffusion & \cellcolor[rgb]{1,0.883,0.8} 0.0431 & \cellcolor[rgb]{1,0.944,0.8} 19.64 & \cellcolor[rgb]{1,0.896,0.8} 0.7767 & \cellcolor[rgb]{1,0.876,0.8} 0.1334 & \cellcolor[rgb]{1,0.800,0.8} 26.42 & \cellcolor[rgb]{1,0.800,0.8} 23.37 \\
RF-Inversion~\cite{rout2025semantic}& flow & \cellcolor[rgb]{1,0.872,0.8} 0.0406 & \cellcolor[rgb]{1,0.907,0.8} 20.82 & \cellcolor[rgb]{1,0.988,0.8} 0.7192 & \cellcolor[rgb]{1,0.969,0.8} 0.1900 & \cellcolor[rgb]{1,0.867,0.8} 25.20 & \cellcolor[rgb]{1,0.889,0.8} 22.11 \\
Ours& VAR & \cellcolor[rgb]{1,0.827,0.8} 0.0305 & \cellcolor[rgb]{1,0.800,0.8} 24.19 & \cellcolor[rgb]{1,0.800,0.8} 0.8370 & \cellcolor[rgb]{1,0.800,0.8} 0.0870 & \cellcolor[rgb]{1,0.855,0.8} 25.42 & \cellcolor[rgb]{1,0.842,0.8} 22.77 \\

\hline
\end{tabular}
\vspace{-0.1cm}
\caption{Quantitative comparison between the proposed method and recent SOTA methods. Our method achieves comparable CLIP similarity to other SOTA methods while demonstrating superior background preservation.}
\vspace{-0.3cm}
\end{table*}

\subsection{Randomness caching}
\label{sec:cache}
The randomness in DF/RF models primarily arises from the initial noise, which often necessitates inversion in training-free editing methods. In contrast, AR models derive their randomness from sampling each token based on its predicted distribution. In contrast to time-consuming inversion techniques~\cite{mokady2023null}, we propose a straightforward yet effective method to ``cache the randomness'' of AR models. This approach allows all necessary information to be stored for future editing use in a single feedforward pass.

Specifically, the encoder $\mathcal{E}$ encodes the input image $\I$ to predict bit labels ${\R}_{queue}$ and transformer inputs $\tilde{\F}_{queue}$ at each level. Given $\tilde{\F}_{queue}$, the transformer $\mathcal{T}$ computes the probability distribution $\P_{k+1} \in \mathbb{R}^{h \times w \times d \times 2}$ for bit labels at each scale using $\tilde{\F}_k$ as input, where $h, w = h_{k+1}, w_{k+1}$, and $d$ is the vocabulary dimension. During inference, both the probability distribution $\P_{queue}$ and selected bit labels ${\R}_{queue}$ are stored. Additionally, the attention map weights $\mathbf{W_k}$ can be cached to facilitate fine-grained control during attention operations, as shown in Algorithm~\ref{alg:caching}.

\begin{algorithm}[t]
\caption{Text-guided image editing}
\begin{algorithmic}[1]
\Require Cached $\R_{queue}$, $\P_{queue}$, $\W_{queue}$, and text embeddings $\Phi(t_T)$ from the target prompt $t_T$, transformer model $\mathcal{T}$, and decoder model $\mathcal{D}$.
\State $\F_0 = \mathbf{0}$
\For{$k = 1, \dots, \gamma$}
\State $\R_k$ = \Call{Q\_Pop}{$\R_{queue}$} \myComment{Q\_Pop:= Queue\_Pop}
\State \Call{Q\_Pop}{${\P}_{queue}$}; \Call{Q\_Pop}{${\W}_{queue}$}
\State $\F_{k}=\F_{k-1}+\up(\R_k, (h, w))$
\EndFor
\For{$k = \gamma+1, \dots, K$}
\State $\P_k, \R_k$ = \Call{Q\_Pop}
{${\P}_{queue}$}, \Call{Q\_Pop}{$\R_{queue}$}
\State $\W_k$ = \Call{Q\_Pop}{$\W_{queue}$} \myComment{Cached attention maps}
\State $\tilde{\F}_{k-1} = \text{down}(\F_{k-1}, (h_k, w_k))$ 
\State $\P_{k}^{tgt} = \mathcal{T}(\tilde{\F}_{k-1}, \Phi(t_T), \W_k)$ \myComment{[Optional] Use \( \W_k \) for attention control in the transformer}
\State $\R_k' \sim \P_k^{tgt}$ \myComment{Random sampling from $\P_k^{tgt}$}
\State $\M_k = \left[(\P_k[..., \R_{k}] - \P_k^{tgt}[..., \R_{k}]) > \tau\right]$
\State $\R^{tgt}_k = \M^k \odot \R'_k + (1-\M^{k}) \odot \R_k$
\State $\F_{k}=\F_{k-1}+\up(\R_k^{tgt}, (h, w))$
\EndFor
\State $\I_t = \mathcal{D}(\F_k)$ 
\State \textbf{Output:} edited image $\I_t$
\end{algorithmic}
\label{alg:edit}
\end{algorithm}

\subsection{Text-guided image editing}
We introduce an additional hyper-parameter $\gamma$ to control the balance between fidelity and generative capabilities. Specifically, for steps where $k \leq \gamma$, we reuse the cached bit labels, thereby preserving low-frequency features, \eg, the overall layout. When $k > \gamma$, the bit labels to be changed are determined using the following adaptive masks. 

\noindent\textbf{Adaptive fine-grained editing mask:} 
To improve the fidelity of edited results, we propose a fine-grained mask \( \M_k \in \mathbb{R}^{h_k\times w_k \times d} \) that adaptively identifies bit labels likely to change. Our approach leverages the distribution difference between the cached probability \( \P_k \) and the newly predicted probability under the target prompt \( t_T \) as an indicator of necessary modifications. Specifically, the fine-grained mask is computed as follows:
\begin{equation}
    \M_k = \left[(\P_k[..., \R_{k}] - \P_k^{tgt}[..., \R_{k}]) > \tau\right],
\end{equation}
where the notation \([..., \R_k]\) denotes the gather operation based on the cached index \(\R_k\), and \(\tau\) is a hyperparameter that balances fidelity and creativity. \(\P_k\) represents the cached probability, as described in Algorithm~\ref{alg:caching}. The target distribution $\P_k^{tgt}$ is predicted by the transformer $\mathcal{T}(\tilde{\F}_{k-1}, \Phi(t_T))$ using the context $\C_k$ as follows:
\begin{equation}
\C_k = (\R_1, ..., \R_\gamma, \R^{tgt}_{\gamma+1}..., \R^{tgt}_{k-1}, \Psi(t_T)).
\end{equation}
This formulation ensures that both the preserved low-frequency structure and the new target prompt contribute to generating an edited image that maintains high fidelity while adhering to the intended modifications.

The rationale behind this design is to consider editing the originally selected bit label only if it experiences a significant probability drop, \eg, having a high probability initially that decreases when the target prompt is used.


\noindent\textbf{Attention Control:}
During the randomness caching phase, by saving the attention maps during the attention operations, we can further involve more fine-grained attention controls~\cite{hertz2022prompt, xu2023inversion} into our editing framework. In this work, we explore an attention refinement strategy that preserve the attention maps from common tokens. The attention map after editing $Edit(\W_k, \W_k^{tgt})$ is defined as follows
\begin{equation}
    (Edit(\W_k, \W_k^{tgt}))_{i,j}:= \begin{cases} 
(\W_k^{tgt})_{i,j} & \text{if } A(j) = \text{None} \\
(\W_t)_{i,A(j)} & \text{otherwise,}
\end{cases}
\end{equation}
where $A$ is the alignment function~\cite{hertz2022prompt} that maps the index of a token in the target prompt to the source prompt, $\W_k^{tgt}$ is the attention map during the editing phase and $\W_k$ is the previously cached one. We only apply attention refinement in the early steps of the generation process, specifically when the spatial resolution of the latent variable is $\leq16$, to ensure finer details in the edited image.

\noindent\textbf{Token re-assembling:}
After obtaining the fine-grained mask $\M$, we use it to guide token sampling during the editing phase from step $\gamma$:  
\begin{equation}  
    \R^{tgt}_k = \M^k \odot \R'_t + (1-\M^{k}) \odot \R_k,  
\end{equation}  
where $\R'_k$ represents bit labels randomly sampled from the distribution $\P_k^{tgt}$. While more advanced source-image-guided sampling strategies for $\R'_k$ could be applied, they typically introduce additional hyper-parameters. Moreover, we find that using the original sampling function for $\R'_k$ already achieves satisfactory results credit to the accuracy of the proposed  adaptive fine-grained mask. Therefore, we keep the sampling of $\R'_k$ unchanged.

\noindent\textbf{Overall:} Our framework is simple while effective, finishing the editing in only two feedforwrad process. The first feedforward go through the given image using our cache mechanism as illustrated in Sec.~\ref{sec:cache}. Then, for the steps with $k>\gamma$, we use the obtained fine-grained mask to re-assemble the bit labels from cached one and edited one. The overall pipeline is illustrated in Algorithm~\ref{alg:edit} and Fig.~\ref{fig:framework}.

\begin{figure*}[t]
    \centering
    \vspace{-0.1cm}
    \includegraphics[width=\linewidth, clip, trim=0 0 0 0]{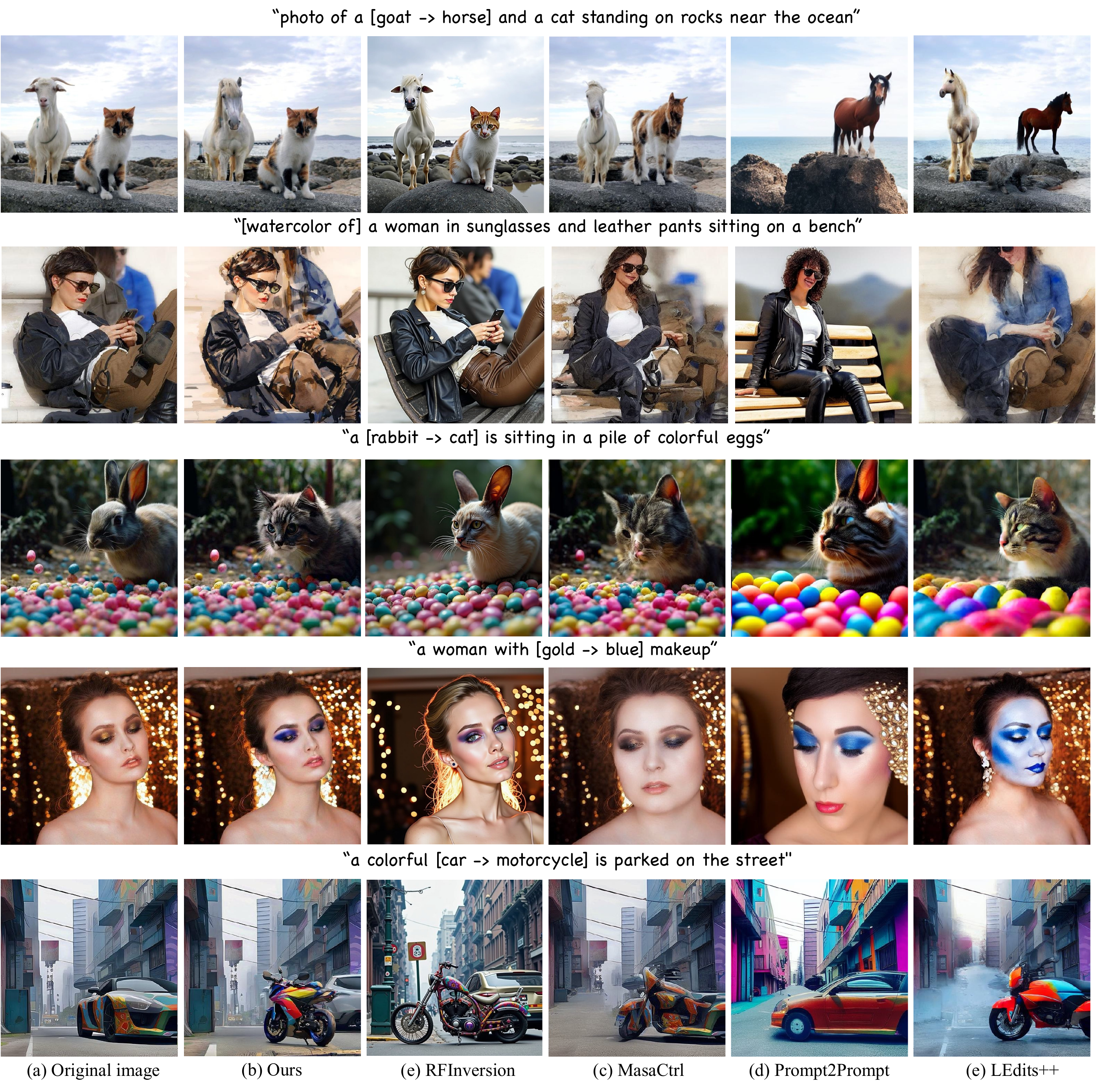}
    \vspace{-0.65cm}
    \caption{Visual comparisons of text-guided image editing results from AREdit~(Ours), RFInversion \cite{rout2025semantic}, MasaCtrl \cite{cao2023masactrl}, Prompt2Prompt \cite{hertz2022prompt}, and LEdits++ \cite{brack2024ledits}. The original image and the source/target prompts or editing instructions are provided. Benefiting from the design of the proposed method and the nature of visual autoregressive models, the proposed method achieves superior performance in detail preservation in the editing-unrelated areas and exhibits a strong ability to follow instructions.}
    \vspace{-0.45cm}
    \label{fig:enter-label}
\end{figure*}

%% file: sec/4_exp.tex
\begin{figure}[t]
    \centering
    \includegraphics[width=\linewidth]{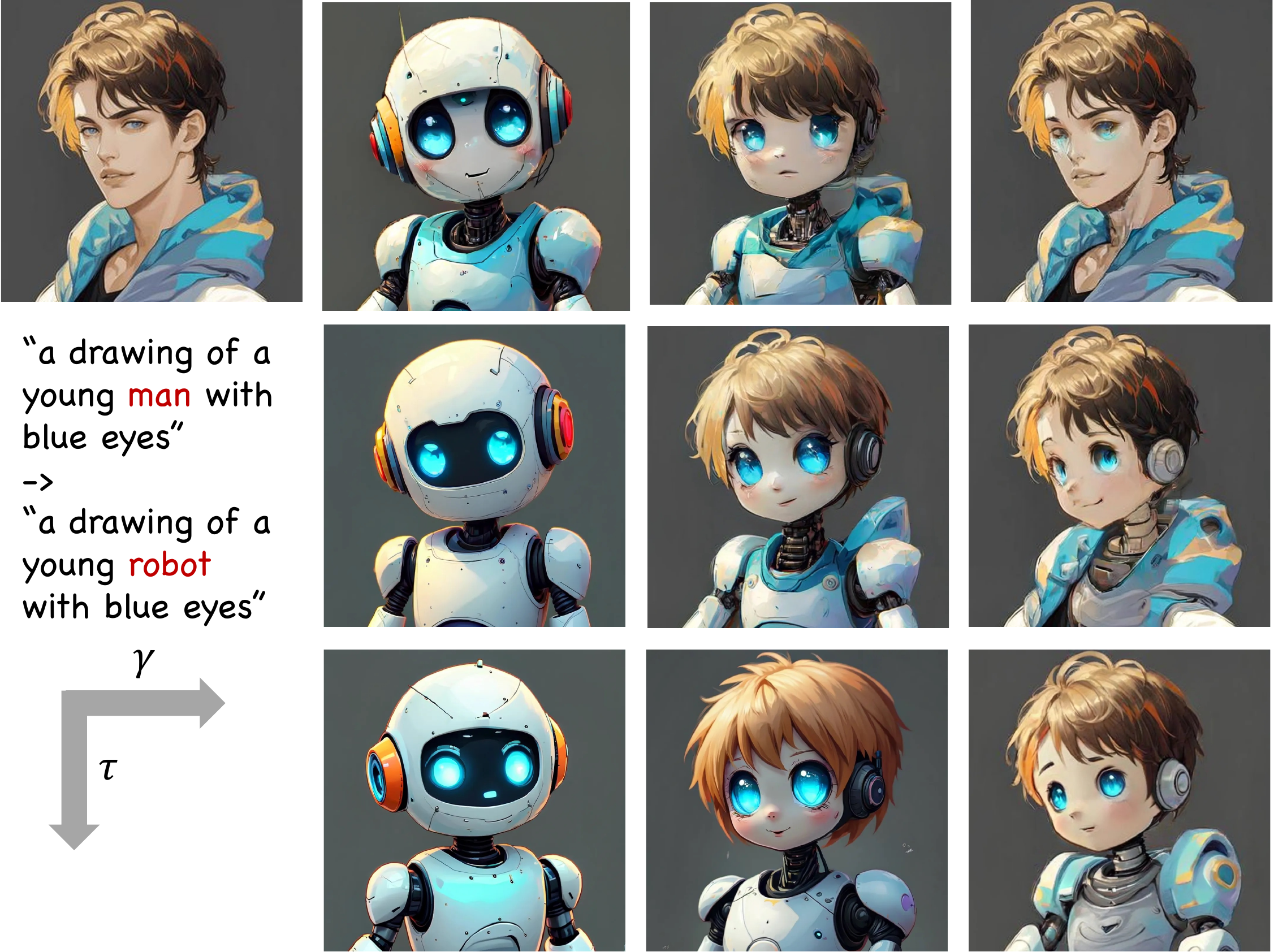}
    \caption{An ablation study to show the effects of our two hyperparameters, $\gamma$ and $\tau$, varying from $1$ to $3$ and $0.4$ to $0.2$ , respectively. The hyperparameter $\gamma$ primarily governs the preservation of low-frequency details, while $\tau$ controls the preservation of high-frequency details, given certain low-frequency information, \ie, a specific value of $\gamma$.}
    \vspace{-0.2cm}
    \label{fig:ablation}
\end{figure}


\section{Experiments and discussions}

\subsection{Experimental settings}
\noindent\textbf{Datasets.} 
We evaluate our method on PIE-Bench~\cite{ju2023direct}, a benchmark comprising 700 images, including both real and artificial visuals. It covers 10 edit types, such as object addition, removal, modification, attribute editing, pose adjustments, style/material transfer, color changes, and background alterations. Each image is paired with an editing mask that delineates the intended modification region and background. Additionally, the benchmark provides an indication of the primary editing subject, which some prior methods leverage for improved fidelity. In contrast, our approach relies solely on source and target prompts to minimize user interaction costs.

\noindent\textbf{Evaluation metrics.} We quantitatively evaluate performance across multiple aspects, including overall structural similarity, text alignment, and preservation of non-edited regions. Text-image consistency is assessed using a CLIP similarity~\cite{wu2021godiva} to measure alignment with the guiding text. To ensure fidelity in unchanged areas, we employ several metrics, including LPIPS~\cite{zhang2018unreasonable}, SSIM~\cite{wang2004image}, and PSNR. Structural distance~\cite{tumanyan2022splicing} is also used to evaluate the overall structural similarity by measuring self-similarity of deep spatial features from DINO-ViT~\cite{caron2021emerging}.

\noindent\noindent\textbf{Implementation details.} We use the latest VAR-based text-to-image foundation model, Infinity-2B~\cite{Infinity}, as our base model. Similar to previous works that employ various attention controllers—such as local blending and attention reweighting based on the primary editing subject—we incorporate the proposed attention controller for certain editing tasks, \ie, modifying background, style, and object color. By default, we set $\gamma=3$ and $\tau=0.2$ for all experiments. When the attention controller is applied, we adjust $\gamma=0$ and $\tau=0.1$ to allow for greater flexibility.
All experiments are conducted on a single Nvidia A100 GPU.

\subsection{Comparisons with previous works}
We select state-of-the-art text-guided image editing approaches as competing methods, including six diffusion-based methods— Prompt2Prompt~\cite{hertz2022prompt}, Pix2Pix-Zero~\cite{parmar2023zero}, MasaCtrl~\cite{cao2023masactrl}, PnP~\cite{tumanyan2023plug}, PnP-DirInv.~\cite{ju2023direct}, and LEdits++ \cite{brack2024ledits}—as well as one rectified flow-based method, RFInversion \cite{rout2025semantic}. In all experiments, we use the official implementations of these methods and report both quantitative and qualitative results with fixed hyper-parameters.

Table~\ref{fig:enter-label} presents the quantitative results, demonstrating that our approach effectively preserves the original content while ensuring edits closely align with the intended modifications.
Fig.~\ref{fig:enter-label} displays the editing results from our method, compared to existing methods. \textit{More visual results can be found in the supplementary.}
Our method effectively handles complex scene editing, such as multi-object scene editing. For example, as shown in the first row of Fig.~\ref{fig:enter-label}, most of the comparison methods confused the modification of the cat and the sheep, either mistakenly changing the cat into the target horse or failing to change the sheep into the horse.
Preserving fidelity is a key challenge in local attribute editing, such as maintaining face identity after modifications. Existing methods often struggle with this, as seen in the fourth row of Fig.~\ref{fig:enter-label}, where face identity is not preserved. In contrast, our approach successfully retains both identity and the intended attribute edits. Additionally, our method effectively handles style transformations, as demonstrated in the second row, preserving the original structure while achieving the target style.



\begin{table}[t]
    \centering
    \scalebox{0.95}{
    \begin{tabular}{llcc}
    \toprule
    Method &  Backbone & Resolution & Speed\\
    \midrule
    LEdits++~\cite{brack2024ledits} & SDXL~\cite{podell2023sdxl} & 1K & 19s (12s) \\
    RFInversion~\cite{rout2025semantic}  & Flux~\cite{flux2024} & 1K & 27s (13.5s) \\
    Ours & Infinity~\cite{Infinity} & 1K & \textbf{2.5s (1.2s)}\\
    \bottomrule
    \end{tabular}}
    \caption{A comparison of computational efficiency across different methods, evaluated on an A100 GPU. Our method demonstrates significantly faster performance compared to other diffusion and rectified flow-based methods. (The values in the bracket indicate the time for subsequent runs without caching or re-inversion.)}
    \label{tab:efficient}
\end{table}
\vspace{-0.2cm}
\begin{figure}[t]
    \centering
    \includegraphics[width=0.95\linewidth]{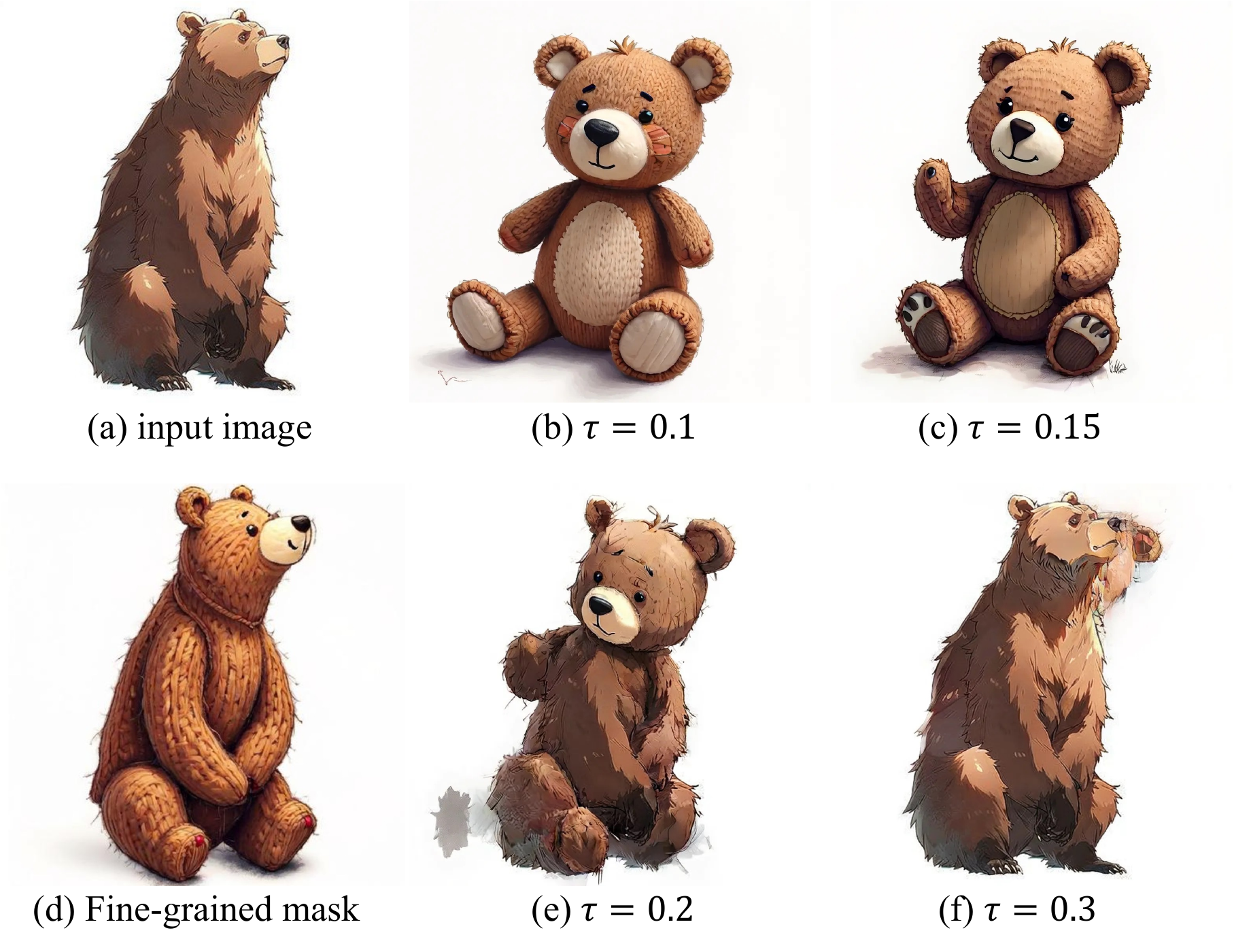}
    \vspace{-0.15cm}
    \caption{Comparison between the proposed adaptive fine-grained mask (d) and the spatial-wise mask (b, c, e, f) with different hyper-parameters. The \(\mathbb{B}^{h_k \times w_k \times d}\) adaptive fine-grained mask enables more precise control over the edited image and preserves more information compared to the spatial-wise mask \(\mathbb{B}^{h_k \times w_k}\). Editing using the spatial-wise mask fails in the cases like style transfers.}
    \vspace{-0.15cm}
    \label{fig:ab_mask}
\end{figure}

\subsection{Ablations and analysis}
\noindent\textbf{Efficiency.}
We assessed and compared the efficiency of various methods using an A100 GPU. Each evaluation was conducted ten times, and we report the average time for a single editing operation in Table~\ref{tab:efficient}, excluding I/O time. Our method demonstrates significantly faster speeds compared to existing techniques based on diffusion models and rectified flows. Additionally, our approach eliminates the need to rerun the caching process when changing hyper-parameters or target prompts, further reducing the editing time to only $\sim$1.2 seconds.

\begin{figure}[t]
    \centering
    \includegraphics[width=\linewidth, trim=0 5 0 0, clip]{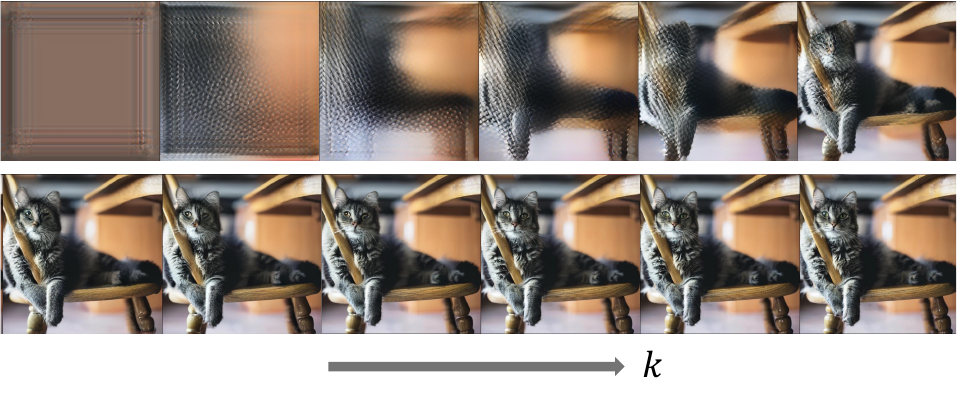}
    \vspace{-0.5cm}
    \caption{The decoded results $\mathcal{D}(\F_k)$ evolve as $k$ increases during the generation process. The early stages primarily contain low-frequency information, such as the overall structure, which can thus be preserved during editing.}
    \vspace{-0.3cm}
    \label{fig:decode_vs_k}
\end{figure}

\noindent\textbf{Effects of hyper-parameters.}
The proposed method involves two hyper-parameters, $\tau$ and $\gamma$, which control the reuse steps of low-frequency features and the threshold for obtaining fine-grained editing masks, respectively. We visualize the editing results with different hyper-parameter settings in Fig.~\ref{fig:ablation}. Specifically, $\gamma$ determines the number of steps for reusing cached bit labels, thereby controlling the extent of low-frequency injection as shown in Fig.~\ref{fig:decode_vs_k}. As depicted in the figure, the relationship between these two hyper-parameters and the generated results is approximately monotonic and intuitive. This allows users to efficiently balance the diversity and fidelity of the generated content, helping them achieve the desired outcomes.

\noindent\textbf{Fine-grained editing mask.}
We find that extending the mask from the spatial dimensions \(\mathbb{B}^{h_k \times w_k} \) to the channel dimension \(\mathbb{B}^{h_k \times w_k \times d}\) is essential for achieving higher granularity, as illustrated in Fig.~\ref{fig:ab_mask}. This enhancement allows our approach to better preserve the original image structure and information. This demonstrates that relying solely on the user-provided spatial mask is insufficient for many editing tasks, such as style, texture, or color modification. Furthermore, our method can be seamlessly combined with the user-provided mask by multiplying the two masks, offering finer-grained control to the user.

\begin{figure}[t]
    \centering
    \includegraphics[width=\linewidth, clip, trim=0 10 0 0]{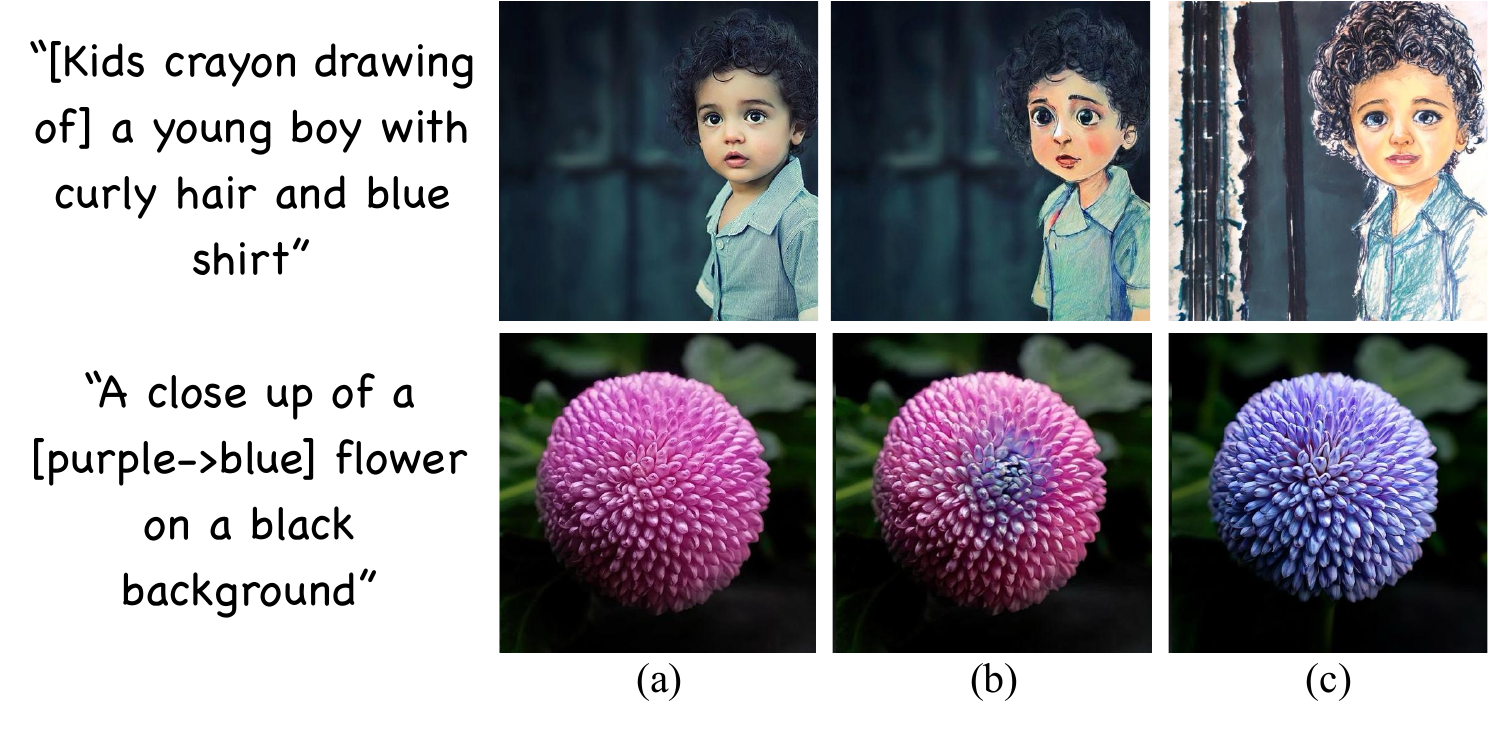}
    \caption{A comparison of results w/ and w/o attention control: (a) Input image, (b) Edited results w/o attention control, and (c) Edited results w/ attention control. By explicitly leveraging the difference between the source and target prompts, \ie, introducing attention control, the proposed method better handles editing tasks involving large modification areas and low-frequency changes.}
    \label{fig:attention_ctrl}
    \vspace{-0.2cm}
\end{figure}

\noindent\textbf{Attention control.}
We find that explicitly utilizing fine-grained control over the cross-attention maps significantly improves performance in editing tasks involving large-area style, attribute, or color changes. Comparisons in Fig.~\ref{fig:attention_ctrl} show that introducing explicit attention control leads to substantial improvements in tasks such as style editing and large-area color modifications. This is because the early stages of autoregressive generation determine the overall color of the entire image, as illustrated in Fig.\ref{fig:decode_vs_k}. To achieve significant modifications over large areas, a small $\gamma$ is required. Since the result at the current scale serves as context for the next scale, directly editing tokens in the early stages poses risks to fidelity. In such cases, attention control serves as an effective supporting mechanism to preserve fidelity.

\begin{figure}[t]
    \centering
    \includegraphics[width=0.9\linewidth]{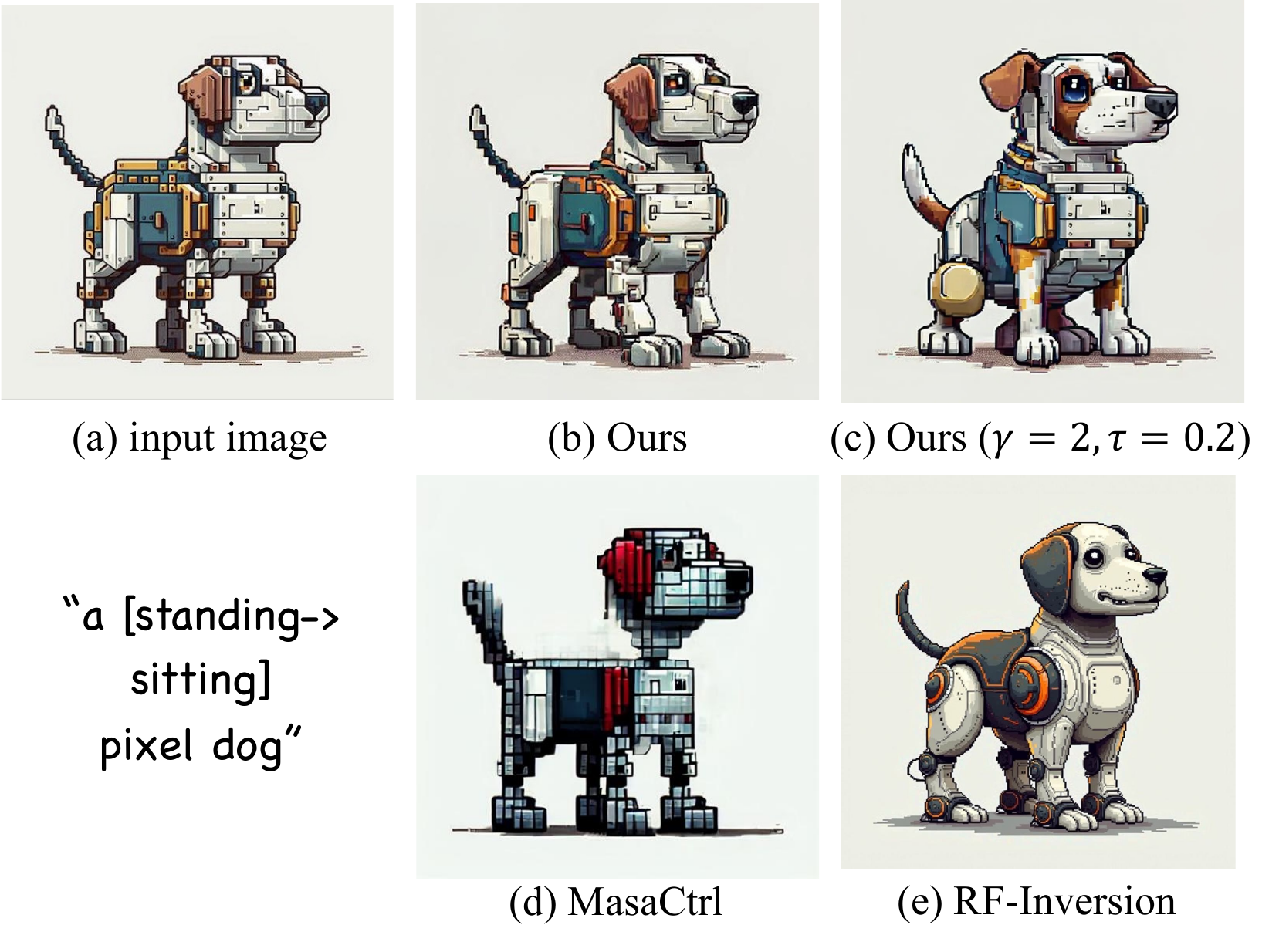}
    \vspace{-0.3cm}
    \caption{A failure case where both the proposed method and other competitors fail. Hyperparameter tuning is required, and/or fidelity may decline in such failure cases.}
    \vspace{-0.4cm}
    \label{fig:failure_case}
\end{figure}

\noindent\textbf{Limitations.}
While the proposed method achieves promising results across various editing tasks, it struggles with certain challenging cases that may exceed the capabilities of the underlying base model. For example, its model size and generative capacity remain weaker than the recently popular RF-based model Flux~\cite{flux2024}. Besides, we find our method also fails in some challenging cases, \eg, Fig.~\ref{fig:failure_case}, where structural rigidity and fine-grained texture details, such as those in robotic features, pose difficulties for discrete token prediction, leading to artifacts and inconsistencies.


%% file: sec/5_conclusion.tex
\section{Conclusion}
We propose a novel VAR-based text-guided image editing framework that effectively addresses the challenges of inversion accuracy and unintended global interactions and modifications present in existing methods. By implementing a caching mechanism to store essential information during the interaction between the source text prompt and image content, we introduce an adaptive fine-grained editing mask that selectively modifies relevant tokens. As a result, our approach ensures precise and controlled editing while preserving the fidelity of unaltered regions. Experiments show that our method achieves SOTA performance, offering superior fidelity and faster inference speeds.